\pdfoutput=1

\documentclass[11pt]{article}

\usepackage[final]{acl}

\usepackage{times}
\usepackage{latexsym}
\usepackage{graphicx} 
\usepackage{multirow}
\usepackage{booktabs}
\usepackage{algorithm}
\usepackage{xcolor}
\usepackage{algpseudocode} 
\usepackage{xcolor}     
\usepackage{tcolorbox}  
\tcbuselibrary{breakable} 

\definecolor{mypromptbg}{rgb}{0.96,0.96,0.92} 
\usepackage[T1]{fontenc}

\usepackage[utf8]{inputenc}

\usepackage{microtype}

\usepackage{inconsolata}

\usepackage{textcomp}
\usepackage{xspace}
\usepackage{tabularx}
\usepackage{url}
\usepackage{footmisc}
\usepackage{subcaption}
\usepackage{amsmath}
\usepackage{enumitem}
\usepackage{listings}          
\usepackage{tcolorbox}
\tcbuselibrary{listings,breakable,skins, most}  
\usepackage{graphicx}

\newcommand{\method}[0]{FLAIRR-TS\xspace}

\usepackage{todonotes}
\usepackage{xcolor}
\usepackage{tcolorbox}
\tcbuselibrary{breakable, skins} 
\usepackage[font=scriptsize,labelfont=bf]{caption} 

\usepackage{listings}
\usepackage{xcolor}
 
\definecolor{codegreen}{rgb}{0,0.6,0}
\definecolor{codegray}{rgb}{0.5,0.5,0.5}
\definecolor{codepurple}{rgb}{0.58,0,0.82}
\definecolor{backcolour}{rgb}{0.95,0.95,0.95}

\lstdefinestyle{mystyle}{
    backgroundcolor=\color{backcolour},   
    commentstyle=\color{codegreen},
    keywordstyle=\color{magenta},
    numberstyle=\tiny\color{codegray},
    stringstyle=\color{codepurple},
    basicstyle=\ttfamily\footnotesize,
    breakatwhitespace=false,         
    breaklines=true,                 
    captionpos=f,                    
    keepspaces=true,                 
    numbers=left,                    
    numbersep=5pt,                  
    showspaces=false,                
    showstringspaces=false,
    showtabs=false,                  
    tabsize=2
}
 
\title{FLAIRR-TS – Forecasting LLM-Agents with Iterative Refinement \\ and Retrieval for Time Series}

\author{Gunjan Jalori \\ Google \\ \texttt{gunjanjalori@google.com} \And
        Preetika Verma \thanks{Work done at Google} \\  Carnegie Mellon University, USA \\ \texttt{preetikv@andrew.cmu.edu} \And
        Sercan Ö Arık \\ Google \\ \texttt{soarik@google.com}} 

\date{}

\begin{document}
\maketitle

\begin{abstract}
Time series Forecasting with large language models (LLMs) requires bridging numerical patterns and natural language. Effective forecasting on LLM often relies on extensive pre-processing and fine-tuning. Recent studies show that a frozen LLM can rival specialized forecasters when supplied with a carefully engineered natural-language prompt, but crafting such a prompt for each task is itself onerous and ad-hoc. We introduce \method, a test-time prompt optimization framework that utilizes an agentic system: a Forecaster-agent generates forecasts using an initial prompt, which is then refined by a refiner agent, informed by past outputs and retrieved analogs. This adaptive prompting generalizes across domains using creative prompt templates and generates high-quality forecasts without intermediate code generation. Experiments on benchmark datasets show improved accuracy over static prompting and retrieval-augmented baselines, approaching the performance of specialized prompts. \method provides a practical alternative to tuning, achieving strong performance via its agentic approach to adaptive prompt refinement and retrieval.
\end{abstract}

\begin{figure*}[t] 
  \centering
  \includegraphics[width=\textwidth]{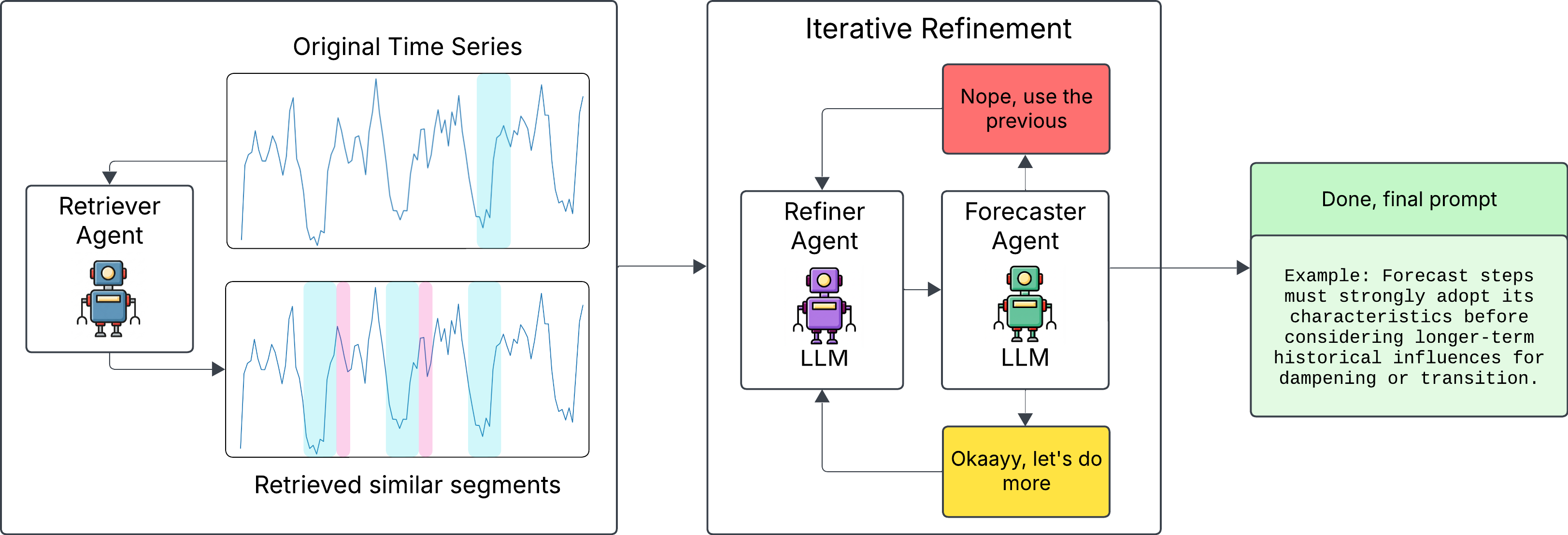} 
  \caption{Flowchart of the the proposed method framework, consisting Retrieval, Forecaster and Refiner agents.} 
  \label{fig:flairr_overview}
\end{figure*}

\section{Introduction}
Recent studies demonstrate that LLMs can leverage their vast pre-trained knowledge to achieve competitive zero-shot and few-shot time-series forecasting (TSF) performance, often rivaling specialized models through direct prompting alone \citep{xue-salim-2023-promptcast-tkde}. The efficacy of LLMs in TSF is often stymied by the \textbf{prompt engineering bottleneck}. The performance of a frozen, pre-trained LLM is critically dependent on the precise natural language prompt it receives. Crafting optimal prompts is currently a laborious, ad-hoc process requiring significant domain expertise and iterative manual tuning for each new dataset or scenario, limiting scalability and robust generalization \citep{niu2024understandingroletextualprompts}. This challenge has spurred research into more sophisticated prompting strategies \citep{liu-zhao-etal-2024-lstprompt, tang-etal-2024-enrichingprompts} and test-time methods without altering weights \citep{jin-etal-2024-timellm}.

Given that LLMs can iteratively refine their outputs through feedback (as in \citet{madaan-etal-2023-selfrefine} and \citet{chen-etal-2025-setsicml}), we explore their capability to autonomously refining their prompts at test time to enhance TSF capabilities.
We introduce \textbf{FLAIRR-TS} - \textbf{F}orecasting \textbf{L}LM-\textbf{A}gents with \textbf{I}terative \textbf{R}efinement and \textbf{R}etrieval, a framework designed to enhance TSF capabilities of LLMs without any training. This approach aims to mitigate the manual prompt engineering burden while simultaneously improving prediction accuracy by grounding forecasts in relevant historical context. FLAIRR-TS integrates a \textbf{Forecaster agent} (F) for initial predictions, a \textbf{Refiner Agent} for Iterative Refinement Tuning (IRT), and a \textbf{Retrieval agent (R)} that sources semantically similar historical time series segments, akin to Retrieval Augmented Generation (RAG) adapted for TSF \citep{han-etal-2023-raft}. This entire cycle of prompt adaptation and forecast refinement occurs without any weight updates, offering a compelling alternative to costly tuning. 
Beyond the capabilities of FLAIRR-TS for general applicability, we also investigate the upper bounds for performance with judiciously-designed prompts. Inspired by \citep{sahoo2025systematicsurveypromptengineering}, we introduce \textbf{Architected Strategy Prompts (ASPs)}, a set of specialized prompts, which include directives for specific analytical procedures or induce particular cognitive approaches. While FLAIRR-TS excels at automated, test-time prompt refinement without prior domain-specific tuning, ASPs allow exploring the performance when, manual strategy-driven design is employed for prompt improvement.

Our main contributions are summarized as:
\begin{itemize}
    \item We propose \textbf{FLAIRR-TS}, a novel prompting and test-time optimization framework for TSF with iterative refinement and retrieval.
    \item We utilize retrieval augmentation for TSF with LLMs with the introduced \textbf{ASPs}, developed via systematic prompt design, to reveal the significant impact of specialized instructions and to serve as high-performance benchmarks.
    \item We demonstrate that FLAIRR-TS consistently improves forecasting accuracy across diverse datasets without model fine-tuning, outperforming static domain agnostic prompting and a non-iterative retrieval-augmented baseline

\end{itemize}


\begin{algorithm*}[htbp] 
\caption{FLAIRR-TS Algorithm}
\label{alg:flairr-ts-concise}
\begin{algorithmic}[1] 

\Require Training data $X$, Historical series $X_{1:t-1}$, Horizon $H$, Initial prompt $P_0$, Context length $L$, \#Segments $M$, Max iterations $N_{\text{iter}}$, Recent ground truth $X_{t:t+H}$, Stopping threshold $\tau_{stop}$
\Ensure Selected prompt $P_{\text{out}}$

\State $P_{\text{curr}} \gets P_0$; \quad $P_{\text{best}} \gets P_0$; \quad $\text{mae}_{\text{min}} \gets \infty$; \quad $\hat{X}_{\text{best}} \gets \text{nil}$; \quad $\text{early\_stop} \gets \text{false}$
\State $X_{\text{HistDB}} \gets X_{1:t-L-1}$; \quad $X_{\text{Ctx}} \gets X_{t-L:t}$ \Comment{Setup context and historical DB}

\For{$k \gets 1$ \textbf{to} $N_{\text{iter}}$}
    \State $S_{\text{retr}} \gets \Call{RetrieveSegments}{X_{\text{HistDB}}, X_{\text{Ctx}}, M}$
    \State $C_{\text{aug}} \gets \Call{AugmentContext}{X_{\text{Ctx}}, S_{\text{retr}}}$
    \State $\hat{X}_{\text{cand}} \gets \Call{ForecasterLLM}{P_{\text{curr}}, C_{\text{aug}}, H}$
    \State $\text{mae}_{\text{curr}} \gets \Call{CalculateMAE}{\hat{X}_{\text{cand}}, X_{t:t+H}}$
    
    \If{$\text{mae}_{\text{curr}} < \text{mae}_{\text{min}}$}
        \State $\text{mae}_{\text{min}} \gets \text{mae}_{\text{curr}}$; \quad $P_{\text{best}} \gets P_{\text{curr}}$; \quad $\hat{X}_{\text{best}} \gets \hat{X}_{\text{cand}}$
    \EndIf
    
    \State $(P_{\text{next}}, \text{done\_signal}) \gets \Call{RefinerLLM}{\{ (P^{(i)}, \text{mae}^{(i)}) \}_{i=0}^{k-1}, P_{\text{curr}}, \tau_{stop}}$ \Comment{Refiner sees history of prompts and errors}
    
    \If{\text{done\_signal}}
        \State $P_{\text{out}} \gets P_{\text{curr}}$; \quad $\text{early\_stop} \gets \text{true}$; \quad \textbf{break}
    \EndIf
    \State $P_{\text{curr}} \gets P_{\text{next}}$
\EndFor

\If{\textbf{not} $\text{early\_stop}$} \Comment{Fallback to best MAE if max iterations reached}
    \State $P_{\text{out}} \gets P_{\text{best}}$
\EndIf

\State \Return $P_{\text{out}}$
\end{algorithmic}
\end{algorithm*}

\section{Methodology}
\subsection{Overall Agentic Architecture}
We propose FLAIRR-TS, a framework combining test-time optimization for iterative refinement via prompting by an agentic system, and retrieval-augmented context to enhance TSF with pre-trained LLMs.
As illustrated in Figure~\ref{fig:flairr_overview} and formally detailed in Algorithm~\ref{alg:flairr-ts-concise}, it operates as a multi-agent system. The \textbf{Forecaster Agent} generates predictions using a prompt that is dynamically improved by the \textbf{Refiner Agent} during an \textbf{Iterative Tuning} phase. This process is enriched by the \textbf{Retrieval Agent} that provides the relevant historical context and augments it to the input provided to the forecaster.
The core iterative cycle (Alg.~\ref{alg:flairr-ts-concise}, lines 7-20) involves forecasting, evaluating the forecast against recent ground truth (e.g., via a metric like MAE), and refining the prompt. The Refiner agent can signal early termination if the forecast improvement falls below a defined threshold, $\tau_{stop}$ (e.g., a 5\% reduction in MAE). Otherwise, if maximum iterations ($N_{\text{iter}}$) are reached, the system defaults to the prompt that yielded the best observed MAE. This adaptive optimization occurs at test-time without any model training.

\subsection{Core Agent Descriptions}

\paragraph{Retrieval Agent.}
Inspired by RAFT \citep{han-etal-2023-raft}, this agent (Alg.~\ref{alg:flairr-ts-concise}, line 8) enhances the Forecaster Agent's inputs by retrieving $M$ historical time series segments ($S_{\text{retr}}$) from a historical database. This database is constructed by applying a sliding window of length $L$ across the entire training split of the dataset. Similarity between the current context window ($X_{\text{Ctx}}$) and the historical segments is measured using \textbf{Pearson's correlation}, with the top-$M$ most similar segments being retrieved. These segments, along with their actual outcomes, provide illustrative examples of past pattern evolutions, directly augmenting the context ($C_{\text{aug}}$) given to the Forecaster-agent (see Appendix~\ref{app:prompt_template} for data formatting details).

\paragraph{Refiner-agent (R).}
Functioning as a meta-optimizer (Alg.~\ref{alg:flairr-ts-concise}, line 12), the Refiner Agent is stateful and analyzes the entire history of the current refinement session. As shown in the algorithm, this history, $\{ (P^{(i)}, \text{mae}^{(i)}) \}_{i=0}^{k-1}$, includes all previously attempted prompts and their resulting errors. By observing this full trajectory, the agent can make non-myopic, informed decisions about the next prompt modification ($P_{\text{next}}$). It provides a $\text{done\_signal}$ if the forecast quality meets the pre-defined termination criterion (i.e., MAE improvement is less than $\tau_{stop}$). Its detailed reasoning, guided by a specific prompt structure (see Appendix~\ref{app:refinerprompt}), might yield feedback such as, \texttt{Pay closer attention to sudden changes in the last 10\% of the input sequence.}

\paragraph{Forecaster-agent (F).}
This agent (Algorithm~\ref{alg:flairr-ts-concise}, line 10) is responsible for generating the time series forecast ($\hat{X}_{\text{cand}}$). It uses the current prompt ($P_{\text{curr}}$)---either the initial prompt $P_0$ or the one refined by the Refiner Agent---along with the augmented context ($C_{\text{aug}}$) provided by the Retriever Agent. FLAIRR-TS allows for the utilization of a potentially more compact LLM as this agent, with its behavior shaped by dynamically optimized prompts. The structure of the prompts is detailed in Appendix~\ref{app:forecasterprompt}.

\subsection{Architected Strategy Prompts (ASP)}
\label{sec:asp}
We present some judiciously-designed prompts inspired by the results of FLAIRR. Instead of merely asking an LLM to predict future values, we aim to induce more complex, imaginative reasoning. If manual prompt engineering is permitted, we aim to improve accuracy further by carefully editing the best prompts from FLAIRR. ASPs are developed by building upon the results achieved by FLAIRR. Some examples include:

\newtcolorbox{aspbox}{breakable,
  colback=mypromptbg,colframe=mypromptbg,
  arc=1mm,boxsep=1.5pt,
  left=2pt,right=2pt,top=1pt,bottom=1pt}

\noindent\textbf{Analytical}\;
\begin{aspbox}
\textbf{Deep STL analysis}: (inspired by \citep{zhou-etal-2024-tempo}) perform an STL decomposition, forecast each component, then recombine them via STL addition.
\end{aspbox}

\vspace{-2pt}
\noindent\textbf{Thinking–Inductive}\;
\begin{aspbox}
\textbf{Monte-Hall Prompting}: frame forecasting as a decision game so the model evaluates several scenarios before committing.
\end{aspbox}

\vspace{-2pt}
\noindent\textbf{Imaginative}\;
\begin{aspbox}
(a) \textbf{Many-Worlds Reasoning}: simulate multiple plausible futures and aggregate them. \hfill

(b) \textbf{D\&D Dungeon-Master}: forecast a character’s hit-point trajectory over upcoming turns.
\end{aspbox}
More details about ASP prompts are in Appendix~\ref{app:prompt-library}.
\begin{table*}[t]
\centering
\resizebox{\linewidth}{!}{%
\begin{tabular}{cc|cccc|cc|cccc}
\hline
\multirow{2}{*}{Dataset} & \multirow{2}{*}{Horizon} & \multicolumn{4}{c|}{Supervised} & \multicolumn{2}{c|}{\begin{tabular}[c]{@{}c@{}}PTMs\end{tabular}} & \multicolumn{4}{c}{\begin{tabular}[c]{@{}c@{}}Prompt\end{tabular}} \\ \cline{3-12}
& & Informer & DLinear & FEDformer & PatchTST & TTM & Time-LLM & LSTP & FLAIRR (Ours) & ASP(G2.5P) (Ours) & ASP(G2.0F) (Ours) \\ \cline{1-12}
\multirow{2}{*}{ETTh1} & 96 & 0.76 & 0.39 & 0.58 & 0.41 & 0.36 & 0.46 & 0.15 & \underline{0.101} & \textbf{0.078} & 0.118 \\
& 192 & 0.78 & 0.41 & 0.64 & 0.49 & 0.39 & 0.54 & \underline{0.22} & 0.246 & \textbf{0.208} & \underline{0.223} \\ \cline{1-12}
\multirow{2}{*}{traffic} & 96  & 0.69 & 0.28 & 0.56 & 0.25 & 0.46 & 0.25    & 0.32    & \underline{0.145} & \textbf{0.143} & 0.184    \\
& 192 & 0.58 & 0.28 & 0.58 & \underline{0.26} & 0.49 & \textbf{0.25}    & 0.31    & 0.326 & 0.324 & 0.296    \\ \cline{1-12}
\end{tabular}
}
\vspace{-0.1in}
\caption{Performance comparison (MAE) of supervised models and zero-shot methods on benchmark datasets. FLAIRR (Ours), ASP(G2.5P) (Ours), and ASP(G2.0F) (Ours) are our proposed/evaluated methods.}
\label{tbl:benchmark_long}
\end{table*}

\begin{table*}[t]
\centering
\vspace{-0.1in}
\small
\setlength{\tabcolsep}{4pt}
\resizebox{0.9\linewidth}{!}{%
\begin{tabular}{cc|cccc|cccc}
\hline
\multirow{2}{*}{Dataset} & \multirow{2}{*}{Horizon} & \multicolumn{4}{c|}{Supervised} & \multicolumn{4}{c}{Prompt} \\ \cline{3-10}
& & Informer & AutoFormer & FedFormer & PatchTST & LSTP & FLAIRR (Ours) & ASP(G2.5P) (Ours) & ASP(G2.0F) (Ours) \\ \hline

\multirow{4}{*}{ILI} 
& 4  & 1.54 & 1.24 & 2.54 & 0.43 & 0.38 & 0.271 & \underline{0.264} & \textbf{0.189} \\
& 12 & 2.33 & 1.82 & 2.67 & 0.43 & 0.39 & 0.249 & \textbf{0.183} & \underline{0.197} \\
& 20 & 2.12 & 1.90 & 1.75 & 1.26 & 0.73 & \underline{0.589} & \textbf{0.564} & 0.867 \\
& 24 & 3.99 & 1.79 & 1.50 & 1.72 & 1.55 & \underline{0.724} & \textbf{0.722} & 1.004 \\ \hline

\multirow{4}{*}{Weather} 
& 24  & 1.45 & 1.38 & 1.95 & 1.55 & 0.17 & \underline{0.110} & \textbf{0.084} & 0.125 \\
& 48  & 1.57 & 1.43 & 1.67 & 1.56 & 0.24 & \underline{0.160} & \textbf{0.142}  & 0.238 \\
& 96  & 1.48 & 1.67 & 1.96 & 1.12 & 0.39 & 0.290 & \underline{0.257}  & \textbf{0.243} \\
& 120 & 1.90 & 1.74 & 2.02 & 1.31 & 0.51 & 0.383 & \textbf{0.309} & \underline{0.369} \\ \hline

\end{tabular}}
\vspace{-0.1in}
\caption{Performance comparison (MAE) on datasets whose test periods post-date the Gemini 2.5 Pro knowledge cut-off. FLAIRR and both ASP variants are ours; \emph{Informer-PatchTST} are supervised baselines; \emph{LSTP} is a prior prompt-based method.}
\label{tbl:benchmark_short}
\end{table*}

\section{Experiments}

Our experiments utilize the Informer \citep{zhou-etal-2021-informer} benchmark datasets\footnote{Full experimental parameters and any dataset-specific preprocessing are in the Appendix.}: ETT (ETTh1, ETTh2, ETTm1, ETTm2), Electricity, and Traffic. We also benchmark on several newer datasets, including Weather and ILINet, and we test on 2025 data to ensure the test period is after the knowledge cutoff date of Gemini. More details are in Appendix~\ref{sec:datasets}. The characteristics of all datasets (domains, frequencies, evaluated horizons $H$) and our approach to data integrity are provided in Section~\ref{sec:datasets}.

\noindent\textbf{LLM Backbone:} We implement FLAIRR using Gemini 2.5 Pro and ASP using both Gemini 2.5 Pro and Gemini 2 Flash. For our ablation study, we replicate these experiments with DeepSeek-V3.

\noindent\textbf{Data \& Execution:} To ensure robust results, we normalize inputs using standard scaling, control for numerical precision in prompts, and report the median performance over five independent runs for each experiment.

\noindent\textbf{Results:} Results are presented in Table~\ref{tbl:benchmark_long} for long-horizon datasets and Table~\ref{tbl:benchmark_short} for short-horizon datasets. We use the Mean Absolute Error (MAE) metric. We compare our work with the most recent prompt-based method, LSTPrompt \citep{liu-zhao-etal-2024-lstprompt} (using a frozen Gemini as its backbone), and two of the best-performing PTMs---TTM \citep{ekambaram2024tinytimemixersttms} and Time-LLM \citep{jin-etal-2024-timellm}. We also compare against non-LLM supervised methods like DLinear \citep{zeng2022transformerseffectivetimeseries}.

\noindent\textbf{Analysis:}
Across 20 distinct scenarios, LAIRR and ASP, outperform all competing models in 14 cases, including every smaller horizon task, and consistently surpass the LSTP baseline on all datasets.

\subsection{Ablations}
\label{sec:ablation}

\begin{figure}[t]  
  \centering
  \includegraphics[width=\columnwidth]{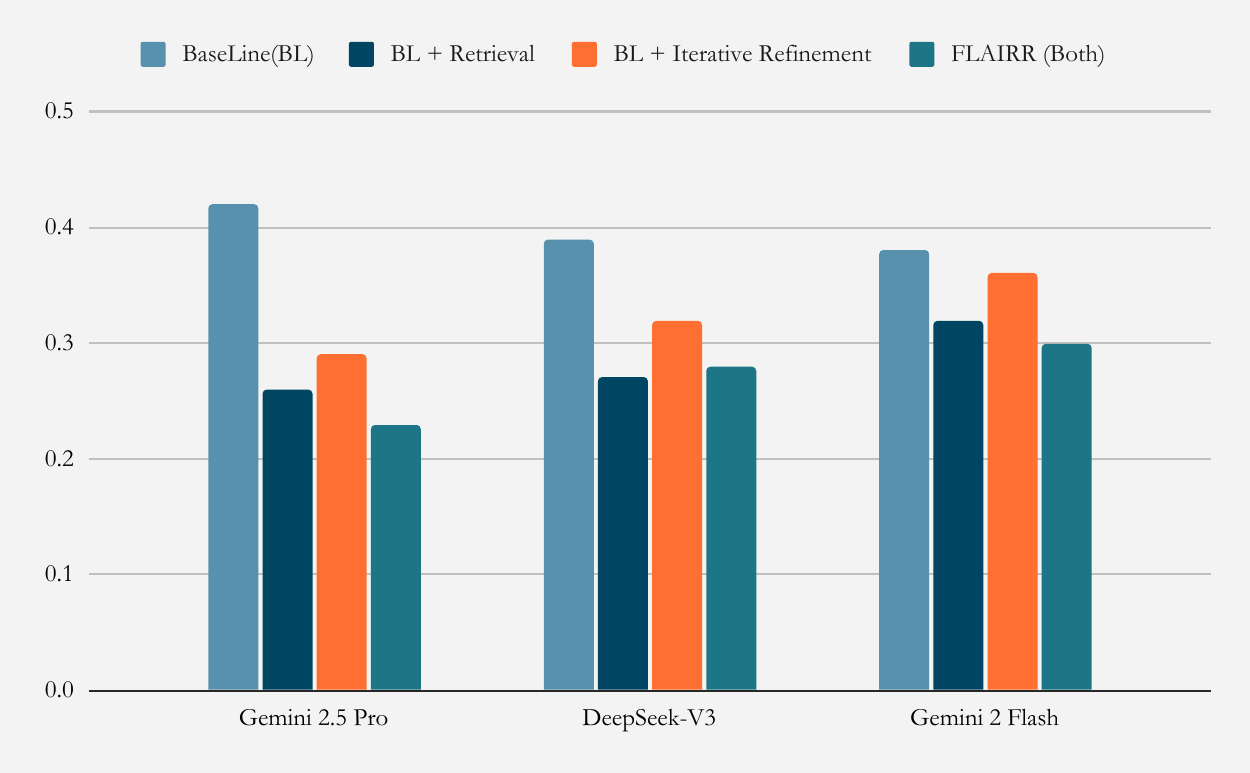}
  \caption{Ablation results, average MAE. Lower MAE is better.}
  \label{fig:ablation}
\end{figure}

\noindent We disentangle the impact of \emph{Retrieval} and \emph{Iterative Refinement (IR)} by successively activating them on top of a \emph{Simple Prompt}. Figure~\ref{fig:ablation} shows the MAE on \textsc{ETTm2} for Gemini 2.5 Pro, Gemini 2 Flash, and the open-source DeepSeek-V3.

\noindent \textbf{Observations:} Retrieval lowers error by grounding forecasts in analogous history, while IR refines outputs through on-the-fly prompt correction. Their combination (\textbf{FLAIRR-TS}) delivers the lowest MAE across all three backbones. Crucially, the same trend holds for DeepSeek-V3, demonstrating that our gains are architecture-agnostic.

\section{Conclusion}

FLAIRR-TS advances the approach to prompting for time-series forecasting. Its core contribution is not to surpass every hand-tuned prompt, but to significantly reduce the burden of manual tuning. The framework provides a systematic and automated process for refining prompts, ensuring consistently high performance from even simple starting instructions. Through its agentic, feedback-driven interactions, FLAIRR-TS offers a scalable pathway to unlocking the full potential of LLMs for forecasting across any dataset or horizon.

\section{Limitations and Future Work}
\begin{itemize}[leftmargin=*]
    \item \textbf{Evaluation coverage.} It'd be important to extend empirical validation to cover robustness to irregular sampling, regime shifts, or domain drifts. 
    \item \textbf{Analogue‐retrieval assumption.} FLAIRR‐TS assumes the presence of semantically similar historical segments. When they do not exist (e.g. for novel events or cold-start prediction scenarios), the refinement loop would come with more risks for compounding errors. 
    \item \textbf{Numerical fidelity of LLMs.} LLMs exhibit limited precision on long or out-of-range sequences, and might hallucinate trends under noise or scale shifts, constraining their reliability. Our method is expected to take advantage of future improvements in LLMs for long input-output modeling and numerical understanding. 
    \item \textbf{Inference cost.} Iterative prompting adds multiple LLM calls per forecast. Latency and energy consumption might be prohibitive for real-time, high-frequency settings, motivating for approaches like LLM distillation.
\end{itemize}

\bibliography{main}

\begin{thebibliography}{18}
\providecommand{\natexlab}[1]{#1}

\bibitem[{Chen and {others}(2025)}]{chen-etal-2025-setsicml}
Zhaofeng Chen and {others}. 2025.
\newblock {SETS}: Self-verification and self-correction for improved test-time scaling.
\newblock Anticipated for International Conference on Machine Learning (ICML).
\newblock Placeholder entry. Details may need updating upon actual publication. Search for preprint by Zhaofeng Chen on SETS.

\bibitem[{Ekambaram et~al.(2024)Ekambaram, Jati, Dayama, Mukherjee, Nguyen, Gifford, Reddy, and Kalagnanam}]{ekambaram2024tinytimemixersttms}
Vijay Ekambaram, Arindam Jati, Pankaj Dayama, Sumanta Mukherjee, Nam~H. Nguyen, Wesley~M. Gifford, Chandra Reddy, and Jayant Kalagnanam. 2024.
\newblock \href {https://arxiv.org/abs/2401.03955} {Tiny time mixers (ttms): Fast pre-trained models for enhanced zero/few-shot forecasting of multivariate time series}.
\newblock \emph{Preprint}, arXiv:2401.03955.

\bibitem[{Gruver et~al.(2023)Gruver, Finzi, Qiu, and Wilson}]{gruver-etal-2023-llmzeroshot}
Nate Gruver, Marc Finzi, Shikai Qiu, and Andrew~Gordon Wilson. 2023.
\newblock \href {https://proceedings.neurips.cc/paper_files/paper/2023/hash/3eb7ca52e8207697361b2c0fb3926511-Abstract-Conference.html} {Large language models are zero-shot time series forecasters}.
\newblock In \emph{Advances in Neural Information Processing Systems 36 (NeurIPS 2023)}, pages 24013--24034.
\newblock ArXiv:2310.07820.

\bibitem[{Han et~al.(2023)Han, Liao, Chiu, Hobbs, An, hwan Oh, Garg, Xiong, and Kim}]{han-etal-2023-raft}
Seungone Han, Peiyuan Liao, Poming~P. Chiu, Jennifer Hobbs, Sungtae An, Min hwan Oh, Vikas~K. Garg, Caiming Xiong, and Yoonkey Kim. 2023.
\newblock \href {https://proceedings.neurips.cc/paper_files/paper/2023/hash/f58465a93999a2appy741e293f849d802-Abstract-Conference.html} {Retrieval augmented time series forecasting}.
\newblock In \emph{Advances in Neural Information Processing Systems 36 (NeurIPS 2023)}, pages 73654--73670.
\newblock ArXiv:2310.16227.

\bibitem[{Jin et~al.(2024)Jin, Wang, Ma, Chu, Zhang, Shi, Chen, Liang, Li, Pan, and Wen}]{jin-etal-2024-timellm}
Ming Jin, Shiyu Wang, Lintao Ma, Zhixuan Chu, James~Y. Zhang, Xiaoming Shi, Pin-Yu Chen, Yuxuan Liang, Yuan-Fang Li, Shirui Pan, and Qingsong Wen. 2024.
\newblock \href {https://openreview.net/forum?id=Unb5CVPtae} {Time-{LLM}: Time series forecasting by reprogramming large language models}.
\newblock In \emph{The Twelfth International Conference on Learning Representations (ICLR)}.
\newblock ArXiv:2310.01728.

\bibitem[{Lewis et~al.(2020)Lewis, Perez, Piktus, Petroni, Karpukhin, Goyal, K{\"{u}}ttler, Lewis, tau Yih, Rockt{\"{a}}schel, Riedel, and Kiela}]{lewis-etal-2020-rag}
Patrick Lewis, Ethan Perez, Aleksandra Piktus, Fabio Petroni, Vladimir Karpukhin, Naman Goyal, Heinrich K{\"{u}}ttler, Mike Lewis, Wen tau Yih, Tim Rockt{\"{a}}schel, Sebastian Riedel, and Douwe Kiela. 2020.
\newblock \href {https://proceedings.neurips.cc/paper/2020/hash/6b493230205f780e1bc26945df7481e5-Abstract.html} {Retrieval-augmented generation for knowledge-intensive {NLP} tasks}.
\newblock In \emph{Advances in Neural Information Processing Systems 33 (NeurIPS 2020)}, pages 9459--9474.

\bibitem[{Liu et~al.(2024)Liu, Zhao, Wang, Kamarthi, and Prakash}]{liu-zhao-etal-2024-lstprompt}
Haoxin Liu, Zhiyuan Zhao, Jindong Wang, Harshavardhan Kamarthi, and B.~Aditya Prakash. 2024.
\newblock \href {https://arxiv.org/abs/2402.16132} {Lstprompt: Large language models as zero-shot time series forecasters by long-short-term prompting}.
\newblock \emph{Preprint}, arXiv:2402.16132.

\bibitem[{Madaan et~al.(2023)Madaan, Tandon, Gupta, Hallinan, Gao, Wiegreffe, Alon, Dziri, Prabhumoye, Yang, Gupta, Majumder, Hermann, Welleck, Yazdanbakhsh, and Clark}]{madaan-etal-2023-selfrefine}
Aman Madaan, Niket Tandon, Prakhar Gupta, Skyler Hallinan, Luyu Gao, Sarah Wiegreffe, Uri Alon, Nouha Dziri, Shrimai Prabhumoye, Yiming Yang, Shashank Gupta, Bodhisattwa~Prasad Majumder, Katherine Hermann, Sean Welleck, Amir Yazdanbakhsh, and Peter Clark. 2023.
\newblock \href {https://arxiv.org/abs/2303.17651} {{Self-Refine}: Iterative refinement with {Self-Feedback}}.
\newblock arXiv preprint arXiv:2303.17651.
\newblock \emph{Preprint}, arXiv:2303.17651.

\bibitem[{Nie et~al.(2023)Nie, Nguyen, Sinthong, and Kalagnanam}]{nie2023timeseriesworth64}
Yuqi Nie, Nam~H. Nguyen, Phanwadee Sinthong, and Jayant Kalagnanam. 2023.
\newblock \href {https://arxiv.org/abs/2211.14730} {A time series is worth 64 words: Long-term forecasting with transformers}.
\newblock \emph{Preprint}, arXiv:2211.14730.

\bibitem[{Niu et~al.(2024)Niu, Zhou, Wang, Sun, and Jin}]{niu2024understandingroletextualprompts}
Peisong Niu, Tian Zhou, Xue Wang, Liang Sun, and Rong Jin. 2024.
\newblock \href {https://arxiv.org/abs/2311.14782} {Understanding the role of textual prompts in llm for time series forecasting: an adapter view}.
\newblock \emph{Preprint}, arXiv:2311.14782.

\bibitem[{Sahoo et~al.(2025)Sahoo, Singh, Saha, Jain, Mondal, and Chadha}]{sahoo2025systematicsurveypromptengineering}
Pranab Sahoo, Ayush~Kumar Singh, Sriparna Saha, Vinija Jain, Samrat Mondal, and Aman Chadha. 2025.
\newblock \href {https://arxiv.org/abs/2402.07927} {A systematic survey of prompt engineering in large language models: Techniques and applications}.
\newblock \emph{Preprint}, arXiv:2402.07927.

\bibitem[{Tang et~al.(2024)Tang, Zhang, Minhas, Li, Chen, Tan, Shah, and Ho}]{tang-etal-2024-enrichingprompts}
Jingyi Tang, Zongyao Zhang, Daksh Minhas, Chengzhang Li, Haomin Chen, Minghuan Tan, Chetan Shah, and Joyce~C. Ho. 2024.
\newblock \href {https://arxiv.org/abs/2407.21368} {Prompting medical large vision-language models to diagnose pathologies by visual question answering}.
\newblock arXiv preprint arXiv:2407.21368.
\newblock \emph{Preprint}, arXiv:2407.21368.
\newblock Verified from.[53] Placeholder `tang-etal-2024-enrichingprompts` resolved.

\bibitem[{Wan et~al.(2024)Wan, Agrawal, Jiang, Choi, and Neubig}]{wan-etal-2024-incontextexemplars}
Yu-Hsiang~Lin Wan, Akshita Agrawal, Chiyu~Max Jiang, Eunsol Choi, and Graham Neubig. 2024.
\newblock \href {https://arxiv.org/abs/2406.18880} {Self-supervised prompting for cross-lingual in-context learning in low-resource languages}.
\newblock arXiv preprint arXiv:2406.18880.
\newblock \emph{Preprint}, arXiv:2406.18880.
\newblock Verified from.[55] Placeholder `wan-etal-2024-incontextexemplars` resolved.

\bibitem[{Xue and Salim(2024)}]{xue-salim-2023-promptcast-tkde}
Hao Xue and Flora~D. Salim. 2024.
\newblock \href {https://doi.org/10.1109/TKDE.2023.3342137} {Promptcast: A new prompt-based learning paradigm for time series forecasting}.
\newblock \emph{IEEE Transactions on Knowledge and Data Engineering}, 36(11):6851--6864.

\bibitem[{Zeng et~al.(2022)Zeng, Chen, Zhang, and Xu}]{zeng2022transformerseffectivetimeseries}
Ailing Zeng, Muxi Chen, Lei Zhang, and Qiang Xu. 2022.
\newblock \href {https://arxiv.org/abs/2205.13504} {Are transformers effective for time series forecasting?}
\newblock \emph{Preprint}, arXiv:2205.13504.

\bibitem[{Zhou et~al.(2021)Zhou, Zhang, Peng, Zhang, Li, Xiong, and Zhang}]{zhou-etal-2021-informer}
Haoyi Zhou, Shanghang Zhang, Jieqi Peng, Shuai Zhang, Jianxin Li, Hui Xiong, and Wancai Zhang. 2021.
\newblock Informer: Beyond efficient transformer for long sequence time-series forecasting.
\newblock In \emph{The Thirty-Fifth {AAAI} Conference on Artificial Intelligence, {AAAI} 2021, Virtual Conference}, volume~35, pages 11106--11115. {AAAI} Press.

\bibitem[{Zhou et~al.(2022)Zhou, Ma, Wen, Wang, Sun, and Jin}]{zhou-etal-2022-fedformer}
Tian Zhou, Ziqing Ma, Qingsong Wen, Xue Wang, Liang Sun, and Rong Jin. 2022.
\newblock \href {https://proceedings.mlr.press/v162/zhou22g.html} {{FEDformer}: Frequency enhanced decomposed transformer for long-term series forecasting}.
\newblock In \emph{Proceedings of the 39th International Conference on Machine Learning (ICML 2022)}, volume 162 of \emph{Proceedings of Machine Learning Research}, pages 27268--27286. PMLR.

\bibitem[{Zhou et~al.(2024)Zhou, Li, Foo, Wang, Soh, Xiong, and Kim}]{zhou-etal-2024-tempo}
Wendi Zhou, Xiao Li, Lin~Geng Foo, Yitan Wang, Harold Soh, Caiming Xiong, and Yoonkey Kim. 2024.
\newblock \href {https://arxiv.org/abs/2405.18384} {{TEMPO}: Temporal representation prompting for large language models in time-series forecasting}.
\newblock arXiv preprint arXiv:2405.18384. Anticipated for NeurIPS 2024.
\newblock \emph{Preprint}, arXiv:2405.18384.

\end{thebibliography}

\clearpage

\appendix

\section{Related Work}
\label{sec:related}
\paragraph{Time Series Forecasting with LLMs:} Traditional time series forecasting has relied on models explicitly trained for the task, from statistical methods to deep architectures like RNN variants and temporal CNNs, up through recent Transformer-based models (e.g. FEDformer \citep{zhou-etal-2022-fedformer} and PatchTST (\citep{nie2023timeseriesworth64})) tailored for long-range sequences. These approaches require substantial training on each target dataset. In contrast, emerging research explores using pre-trained LLMs as general-purpose forecasters via prompting at inference time only, without gradient-based fine-tuning. \citet{xue-salim-2023-promptcast-tkde} pioneered this direction with PromptCast, formulating forecasting as a prompt–completion task: historical values are encoded into a textual prompt (possibly with instructions) and the LLM’s next-token predictions are decoded as forecasts. \citet{gruver-etal-2023-llmzeroshot} similarly represent numerical time series as token sequences and treat extrapolation as language modeling, finding that GPT-3 and LLaMA-2 can zero-shot extrapolate time series with accuracy comparable to or exceeding specialized trained models. TNotably, these LLM-based approaches leverage the models’ strong sequence modeling and few-shot generalization for competitive benchmark results, without requiringabilities to achieve competitive results on standard benchmarks without any task-specific training data. Nevertheless, naive prompt formulations might overlook important temporal dynamics and patterns. Recent works therefore propose more advanced test-time prompting strategies. \citet{liu-zhao-etal-2024-lstprompt} introduce LSTPrompt, which splits the prediction into short- and long-term sub-tasks and guides the LLM through a chain-of-thought reasoning process; this method outperforms earlier prompt baselines and even approaches the accuracy of dedicated TS models. \citet{tang-etal-2024-enrichingprompts} report that enriching prompts with external knowledge (e.g. known seasonal periods or contextual clues) and using natural language rephrasings of the input can significantly improve an LLM’s forecasting accuracy. Another technique, Time-LLM \citep{jin-etal-2024-timellm}, reprograms a frozen LLM by mapping time-series data into textual “patches” and prepending learned prompt tokens, allowing the model to output forecasts that outperform state-of-the-art specialized forecasters without any fine-tuning of the LLM’s weights. On the other hand, \citet{zeng2022transformerseffectivetimeseries} offer a cautionary perspective: through extensive ablations, they found that removing the LLM or replacing it with a simple attention-based network in these pipelines often does not hurt performance (and sometimes improves it), calling into question how much current LLM-for-TS methods truly benefit from the pre-trained language model. To push LLM-based forecasting further, researchers are drawing on insights from prompt optimization and test-time reasoning. For example, \citet{wan-etal-2024-incontextexemplars} show that intelligently selecting and reusing in-context exemplars can yield larger gains than optimizing instructions alone, suggesting that careful few-shot prompt design is crucial. \citet{chen-etal-2025-setsicml} propose a self-verification and self-correction framework (SETS) that lets the model iteratively refine its outputs at inference, achieving better accuracy scaling on complex reasoning tasks. Incorporating such techniques into zero-shot forecasting prompts is an exciting direction. In summary, the literature demonstrates a nascent but growing paradigm of using pre-trained LLMs directly for time series forecasting, with multiple studies showing that, given the right prompts, foundation models can attain forecast accuracy rivaling traditional specialized models. While these methods demonstrate progress in leveraging LLMs for forecasting, the dynamic and optimal design of prompts—especially those needing to integrate complex reasoning, external knowledge, and iterative feedback—remains a key challenge. Our work, FLAIRR-TS, aims to address this by structuring the forecasting process around specialized agents for dynamic prompt adaptation and refinement.

\paragraph{Agentic Frameworks with Iterative Refinement}
The concept of employing multiple interacting agents or distinct processing roles for complex problem-solving has gained traction in AI. Such agentic systems can distribute tasks, specialize functionalities, and enable more sophisticated reasoning or generation processes. Iterative refinement, where an output is progressively improved through feedback loops, is a common characteristic of these systems and is also seen in self-correction mechanisms within single LLMs (e.g., Self-Refine by \citet{madaan-etal-2023-selfrefine}). For instance, systems might involve a generator agent and a critic agent, or distinct agents for planning, execution, and verification. FLAIRR-TS draws inspiration from these paradigms by structuring its operation around specialized agents: a Forecaster-agent for initial prediction, a retriever agent for sourcing relevant context, and a refiner agent for iterative prompt refinement. This agentic decomposition facilitates more targeted and adaptable modifications to distinct aspects of the forecasting prompt through these specialized roles. Crucially, unlike traditional multi-agent systems where agents might be independently trained or involve complex coordination protocols, FLAIRR-TS implements these roles using LLMs at test time to dynamically adapt the prompting strategy itself. The "refinement" occurs in the textual instructions and contextual information fed to the LLM, rather than through updates to model weights, distinguishing it from model distillation or training paradigms. This focus on inference-time prompt adaptation through an agentic perspective is a key aspect of our approach. This structured approach also aims to ensure that the LLM's reasoning and generative capabilities are a core component of the forecasting process, addressing concerns about their actual contribution in some prior LLM-for-TS pipelines.

\paragraph{Retrieval Augmented Generation:}
Retrieval Augmented Generation (RAG) \citep{lewis-etal-2020-rag} has become a standard technique for enhancing LLMs in knowledge-intensive NLP tasks. RAG systems retrieve relevant documents or passages from an external corpus and provide them as additional context to the LLM, improving factual grounding and reducing hallucination. Recently, \citet{han-etal-2023-raft} adapted this concept to time series forecasting with their Retrieval Augmented Time Series Forecasting (RAFT) approach. RAFT retrieves historical time series segments similar to the current input window and uses them to augment the context provided to a forecasting model (in their case, an LLM). Our work directly builds upon and integrates the RAFT principle within the Retrieval agent component of FLAIRR-TS. We hypothesize that the effectiveness of RAFT can be further enhanced by optimizing the prompt that instructs the LLM on how to utilize the retrieved historical context, which is precisely what the agentic interaction within FLAIRR-TS aims to achieve.
\section{Refiner Agent}
\label{app:refinerprompt}
\smallskip

\begin{tcolorbox}[
  breakable,
  enhanced,
  listing options={breaklines=true},
  colback=mypromptbg,
  colframe=mypromptbg,
  arc=1mm,
  boxsep=2pt,
  left=3mm,
  right=3mm,
  top=2mm,
  bottom=2mm
]
\small   

You are an expert Time-Series-Forecasting \textbf{Prompt Engineer} acting as a \textbf{Refiner Agent}.  
Your goal is to analyze a set of forecasting attempts made by a \textbf{Forecaster Agent} and provide specific, actionable \textbf{Learnings} on how to improve the \emph{initial forecasting prompt} used by the Forecaster.  
The Forecaster Agent uses a base prompt and adds new forecasting instructions to it based on your learnings.

\bigskip
\noindent\textbf{Key Information for Your Analysis for this Iteration} \verb|{it + 1}|:
\begin{enumerate}[label=\arabic*., leftmargin=*]
  \item Current Forecasting Instructions Under Review: 
        \verb|{current_instructions_under_review}|
  \item Overall Mean Absolute Error (MAE) for this batch of samples: \verb|{mae_to_report_to_teacher}|
\end{enumerate}

\medskip
You will also be given a batch of individual samples, where each sample includes:
\begin{enumerate}[label=\arabic*., leftmargin=*]
  \item The full \texttt{Prompt} the Forecaster Agent used (includes the instructions above).
  \item The Forecaster Agent's \texttt{Predictions} for the \texttt{OT} variable.
  \item The \texttt{Ground-Truth} \texttt{OT} values.
\end{enumerate}

\medskip
\noindent\textbf{Your Analysis Task:}
\begin{enumerate}[label=\arabic*., leftmargin=*]
\item \textbf{Identify error patterns.} Compare \texttt{Predictions} with \texttt{Ground Truths}.
      Look for systematic errors (e.g., over/under-prediction, lagging, volatility mishandling).
\item \textbf{Correlate errors with prompts and instructions.} 
      Check whether the current instructions are ambiguous, misleading, too complex, or otherwise harmful.
\item \textbf{Formulate "Learnings".} 
      Give concrete, generalizable improvements (e.g., adjust look-back horizon, drop STL decomposition, add weekday feature).
\item \textbf{Determine "Done" status.}
      \begin{itemize}[nosep,leftmargin=*]
        \item If the percentage reduction in MAE is less than the pre-defined stopping threshold ($\tau_{stop}$), output \texttt{Done: True}. 
        \item Otherwise output \texttt{Done: False}.
      \end{itemize}
\end{enumerate}

\medskip
\noindent\textbf{Output Format—exactly this template}

Learnings:
\texttt{<your concise, actionable suggestions here>}

Done: \texttt{<True or False>}

Confidence in output: \texttt{<High | Medium | Low>} – one-line rationale.
\end{tcolorbox}

\section{Forecaster Agent}
\label{app:forecasterprompt}
\medskip
\noindent\textbf{Prompt-Synthesis Instructions}

\smallskip
\noindent\textit{Example: Forecasting-Instruction Refinement}

\begin{tcolorbox}[
  breakable,
  enhanced,
  colback=mypromptbg,
  colframe=mypromptbg,
  arc=1mm,
  boxsep=2pt,
  left=3mm, right=3mm,
  top=2mm, bottom=2mm,
  fontupper=\small
]

\textbf{You are an intelligent agent} that synthesizes forecasting prompts based on expert feedback.  
You will receive \textit{Learnings} from a \textbf{Refiner Agent} that suggest improvements to an initial time-series forecasting prompt.  
Your task is to turn these learnings into concise and effective \emph{prompt-forecasting instructions}.  
These instructions will be appended to a base forecasting prompt.

\medskip
\textbf{The forecasting instructions should:}
\begin{itemize}[leftmargin=*]
  \item Be a short set of guiding principles (maximum 3 actionable items).  
  \item Directly address the issues and suggestions in the Learnings.  
  \item Be clearly phrased for another LLM to follow.  
  \item **Do not include placeholders such as \texttt{\{previous\_data\}} or \texttt{\{prediction\_data\}}.
  \item **Do not change the output format or the forecasting task itself.
  \item If no actionable learnings exist, output a safe generic set—or state:
  
        \textit{No specific new instructions generated due to lack of actionable learnings.}
\end{itemize}

\medskip
\textbf{Example (Refiner Agent said “focus on recent volatility”):}
\begin{description}[leftmargin=*,labelindent=0pt]
  \item[Learnings:] The model often misses sudden spikes; the prompt should ask the forecaster to pay more attention to recent volatility and its effect on the next step.
  \item[Your Output (forecasting instructions):]
        “Critically assess the volatility in the most recent data points. Your forecast for the next step should reflect whether this volatility is expected to continue, increase, or decrease. Explain this assumption in your reasoning.”
\end{description}

\medskip
\textbf{Learnings you received:}\\
\verb|{current_learnings}|

\smallskip
Based on these learnings, generate only the refined prompt-forecasting instructions below (no extra commentary).

\medskip
\textbf{Refined Prompt Forecasting Instructions:}

\texttt{<model prediction here>}
\end{tcolorbox}

\section{Prompt template}
\label{app:prompt_template}

\begin{tcolorbox}[
  breakable,
  enhanced,
  colback=mypromptbg,
  colframe=mypromptbg,
  arc=1mm,
  boxsep=2pt,
  left=3mm,  right=3mm,
  top=2mm,   bottom=2mm,
  fontupper=\small       
]

\textbf{Objective}\par
Provide a well-reasoned forecast for the \verb|{target_variable}| value in the next row of the dataset, given the historical data.

\bigskip
\textbf{Dataset Instructions}
\begin{itemize}[leftmargin=*]
  \item \textbf{Dataset}: \texttt{data\_name}, \texttt{data\_description}
  \item \textbf{Variable to Predict}: \verb|{target_variable}|.
  \item \textbf{Task}: Predict the \verb|{target_variable}| values for the next
        \verb|{prediction_length}| steps using the historical data.
  \item \textbf{Constraints}:
        \begin{itemize}[leftmargin=*]
           \item Adhere strictly to the specified output format.
        \end{itemize}
\end{itemize}

\textit{If} \verb|instructions:| 

\verb|Forecasting Instructions:|  
\verb|{instructions}|

\medskip
\textit{If} \verb|raft_context:| The \texttt{\{raft\_context\}} contains the M retrieved historical segments. Each segment and its corresponding ground-truth outcome are formatted as comma-separated text strings.

\bigskip
\textbf{Input Data}
\begin{itemize}[leftmargin=*]
  \item Historical Data:  
        \verb|{previous_sequence_length_data}|
\end{itemize}

\bigskip
\textbf{Output Format — exactly this}
\begin{verbatim}
Predicted Values: [predicted_value_1, ...]
Reasoning: [Your detailed reasoning ]
Certainty Estimate: [Percentage certainty]
Certainty Reasoning: [reasoning]
\end{verbatim}

\end{tcolorbox}


\tcbset{forecaststyle/.style={
    breakable,
    enhanced,
    colback=mypromptbg,
    colframe=mypromptbg,
    arc=1mm,
    boxsep=2pt,
    left=3mm,right=3mm,
    top=2mm,bottom=2mm,
    fontupper=\ttfamily\small   
}}

\section{Prompt Library}
\label{app:prompt-library}

The following library of prompts was designed to test different cognitive pathways of the LLM. The strategies are grouped into categories: Analytical prompts that enforce a structured decomposition of the problem; Thinking-Inductive prompts that encourage probabilistic or scenario-based reasoning; and Imaginative prompts that use metaphor and creative framing to elicit novel patterns.

\subsubsection*{teacher-student-loop}
\begin{tcolorbox}[forecaststyle]
ACT I — REFINER AGENT
Propose a first-pass forecast for the next \verb|{sequence_length}| steps.

ACT II — FORECASTER AGENT
Evaluate the Refiner's forecast against the most recent known data and
suggest corrections.

ACT III — REFINER AGENT
Incorporate feedback and provide the refined forecast.
\end{tcolorbox}

\subsubsection*{self-verification-sets}
\begin{tcolorbox}[forecaststyle]
Step 1 – Generate candidate forecast A for \verb|{sequence_length}| steps.  
Step 2 – Generate independent candidate forecast B.  
Step 3 – For each horizon \emph{h}, if the two differ beyond an acceptable
tolerance, reconcile them (e.g., by averaging).  
Provide only the reconciled forecast.
\end{tcolorbox}

\subsubsection*{meta-prompt-conf-bands}
\begin{tcolorbox}[forecaststyle]
Forecast \verb|{sequence_length}| steps and include 68\,\% and 95\,\% confidence
bands. Briefly explain the uncertainty assumptions before the numbers.
\end{tcolorbox}

\subsubsection*{imaginary-python-repl}
\begin{tcolorbox}[forecaststyle]
You are \textbf{ForecastPy}, a mental Python REPL.  
Think then “run code in your head’’ that derives the forecast for the next
\verb|{sequence_length}| steps.
\end{tcolorbox}

\subsubsection*{synesthetic-soundtrack}
\begin{tcolorbox}[forecaststyle]
Interpret the past sequence as MIDI velocity (0–127) and compose the
next \verb|{sequence_length}| beats that extend the melody.  
Provide both the MIDI integers and the values rescaled to original units.
\end{tcolorbox}

\subsubsection*{color-gradient-canvas}
\begin{tcolorbox}[forecaststyle]
Map each value to an RGB triplet on a blue-to-red gradient.  
Produce a grid of HEX colours that encodes the next
\verb|{sequence_length}| points.
\end{tcolorbox}

\subsubsection*{dungeon-master}
\begin{tcolorbox}[forecaststyle]
You are a D\&D Dungeon Master. The party's HP over the last turns is
shown. Forecast HP for the next \verb|{sequence_length}| turns, assuming no boss
fights and only mild potion use.
\end{tcolorbox}

\subsubsection*{micro-essay-poisson}
\begin{tcolorbox}[forecaststyle]
Write a $\le$60-word micro-abstract describing the generative mechanism,  
then list \verb|{sequence_length}| $\lambda$ parameters for a Poisson baseline.
\end{tcolorbox}

\subsubsection*{reverse-sudoku}
\begin{tcolorbox}[forecaststyle]
Think of the next \verb|{sequence_length}| points as filling a
$9\times11$ Sudoku-like grid whose row sums match the recent history.  
Provide the grid and a flattened list.
\end{tcolorbox}

\subsubsection*{many-worlds-ensemble}
\begin{tcolorbox}[forecaststyle]
Create forecasts for four parallel universes (A–D) shifted by $-2\sigma$,
$-1\sigma$, $+1\sigma$, $+2\sigma$, each \verb|{sequence_length}| steps long,  
then provide a consensus median forecast.
\end{tcolorbox}

\subsubsection*{haiku-seeded}
\begin{tcolorbox}[forecaststyle]
Compose a three-line haiku that metaphorically describes the upcoming
pattern, then list the \verb|{sequence_length}| numeric forecasts,
one per line.
\end{tcolorbox}

\section{Datasets}\label{sec:datasets}
Experiments were performed on a diverse set of widely-used time-series-forecasting (TSF) benchmark datasets spanning multiple domains, sampling frequencies, and statistical characteristics (e.g., seasonality, trend, noise levels).  All datasets are normalized with StandardScaling from sklearn package.
The datasets are:

\begin{itemize}[leftmargin=*]
    \item \textbf{ETT} (\texttt{ETTh1}, \texttt{ETTh2}, \texttt{ETTm1}, \texttt{ETTm2}) – Electricity Transformer Temperature data recorded at hourly (\texttt{h}) or 15-minute (\texttt{m}) intervals; widely used for long-sequence forecasting with OT as target variable (ETTh: 17,420 total data points, ETTm: 69,680 total data points)
    \item \textbf{Electricity} – Hourly household electricity data of customers with electricity consumption as target variable (26,304 total data points)
    \item \textbf{Traffic} – Hourly occupancy rates from California road-traffic sensors (2021-2025 March) with traffic volume as target variable (17,544 data points)
    \item \textbf{ILINet} – Weekly Influenza-Like-Illness counts from the CDC (2002-2025 April) with total ILI patients as target variable  (1,441 total data points)\footnote{https://gis.cdc.gov/grasp/fluview/fluportaldashboard.html}
    \item \textbf{Weather} - Hourly weather data from Chicago with temperature as target variable (35,052 total data points)\footnote{https://www.kaggle.com/datasets/curiel/chicago-weather-database}
\end{itemize}

\subsection{Data Integrity}
A significant consideration when utilizing Large Language Models (LLMs) for time series forecasting is the potential for the model's pre-training data to inadvertently include samples from the test set, which could lead to an overestimation of predictive performance. To rigorously uphold data integrity in this study, we employed ILINet and weather datasets as benchmarks, with a specific focus on temporal data separation. Our experimental design ensures that all data samples within the test set originate from dates strictly subsequent to the known training data cut-off date of the LLM employed for inference. This chronological separation mitigates the risk of test data contamination, providing a robust and fair evaluation of the LLM's ability to generalize and forecast genuinely unseen future values.


\subsection{Evaluation Metrics}\label{sec:metrics}
Forecasting performance was assessed with two standard error metrics:

\begin{align}
\mathrm{MAE} &= \frac{1}{H}\sum_{i=1}^{H}\left|\hat{X}_{t+i} - X_{t+i}\right|,
\end{align}
Where $H$ is the prediction horizon, $\hat{X}_{t+i}$ is the predicted value, and $X_{t+i}$ is the ground-truth value.  
Lower values indicate better performance for both metrics.  
These metrics were computed directly from the experimental results.


\subsection{Hyperparameter Settings}
\label{sec:appendix_hyperparams}

Here we detail the key hyperparameters used for the FLAIRR-TS framework in our experiments.

\begin{description}
    \item[\texttt{ts\_max\_iter}] \textbf{Value:} 5. The maximum number of refinement iterations allowed in the iterative refinement loop.

    \item[\texttt{ts\_stopping\_criteria}] \textbf{Value:} 5\%. The refinement process is stopped early if the percentage reduction in Mean Absolute Error (MAE) between iterations falls below this threshold.

    \item[\texttt{ts\_sample\_size}] \textbf{Value:} 3. The number of validation samples used within each refinement iteration to evaluate the quality of a candidate prompt.

    \item[\texttt{raft\_m\_retrieval}] \textbf{Value:} 2. The number of similar historical segments retrieved by the Retrieval Agent to be used as context.
\end{description}
\section{Future directions}

There are several avenues for future work. One direction is to incorporate quantitative validation in the loop: currently, the Refiner-agent's feedback quality is not directly measured. If we had a small hold-out set or could use the model’s own likelihood of the data, we might select or weight feedback. This leans towards techniques in automatic prompt optimization where a reward is defined. Additionally, while FLAIRR-TS currently uses natural language for feedback from the Refiner-agent, one could imagine hybrid approaches where the Refiner-agent suggests pseudo-code or formulaic adjustments (if the LLM agents are equipped with a calculator tool). That could improve handling of scale and magnitude issues. On the retrieval side, exploring more advanced analog search (perhaps using learned embeddings or matching not just on raw values but pattern descriptors) might yield even more relevant cases to show the Refiner-agent, especially for complex multivariate data.

From an application perspective, deploying FLAIRR-TS in an interactive forecasting system would be very interesting. Because FLAIRR-TS’s intermediate steps (the prompts, the retrieved analogs, the feedback) are human-readable, a human analyst could intervene in the loop – agreeing or disagreeing with the Refiner-agent's critique, or adding their own feedback. This could turn forecasting into a collaborative dialog between human, Forecaster-agent, and Refiner-agent. In settings like supply chain or epidemiology forecasting, such a system could help build trust as well, since each refinement step can be scrutinized.

\end{document}